\pdfoutput=1

\documentclass[11pt]{article}

\usepackage{EMNLP2022}

\usepackage{times}
\usepackage{latexsym}

\usepackage[T1]{fontenc}

\usepackage[utf8]{inputenc}

\usepackage{microtype}

\usepackage{inconsolata}
\usepackage{multirow, makecell}
\usepackage{amsmath}
\usepackage{amsfonts} 
\usepackage{caption}
\usepackage{subcaption}
\usepackage{graphicx}

\title{Are Neighbors Enough? \\ Multi-Head Neural $n$-gram can be Alternative to Self-attention}
\author{Mengsay Loem, Sho Takase, Masahiro Kaneko, \and Naoaki Okazaki \\
  Tokyo Institute of Technology \\ 
  \texttt{\{mengsay.loem, sho.takase, masahiro.kaneko\}@nlp.c.titech.ac.jp} \\\texttt{naoaki.okazaki@c.titech.ac.jp}}

\begin{document}
\maketitle
\begin{abstract}
Impressive performance of Transformer has been attributed to self-attention, where dependencies between entire input in a sequence are considered at every position.
In this work, we reform the neural $n$-gram model, which focuses on only several surrounding representations of each position, with the multi-head mechanism as in \newcite{transformer-vaswani}.
Through experiments on sequence-to-sequence tasks, we show that replacing self-attention in Transformer with multi-head neural $n$-gram can achieve comparable or better performance than Transformer.
From various analyses on our proposed method, we find that multi-head neural $n$-gram is complementary to self-attention, and their combinations can further improve performance of vanilla Transformer.
\end{abstract}

\section{Introduction}
\label{sec:intro}

Since \newcite{transformer-vaswani} proposed Transformer, it and its variants have been applied to various tasks such as machine translation, summarization, and automatic speech recognition~\cite{transformer-vaswani,takase-okazaki-2019-positional,speech-transformer}.
The important module of Transformers is the self-attention which computes the representation at the focused position based on dependencies between an input representation at the focused position and all input representations.
In more detail, for a sequence of input representations $x_1, ..., x_L$, we compute the output representation at the position $t$ based on the attention~\cite{bahdanau:2015:ICLR} between $x_t$ and $x_1, ..., x_L$.

\begin{figure}[!t]
  \centering 
  \includegraphics[width=8cm]{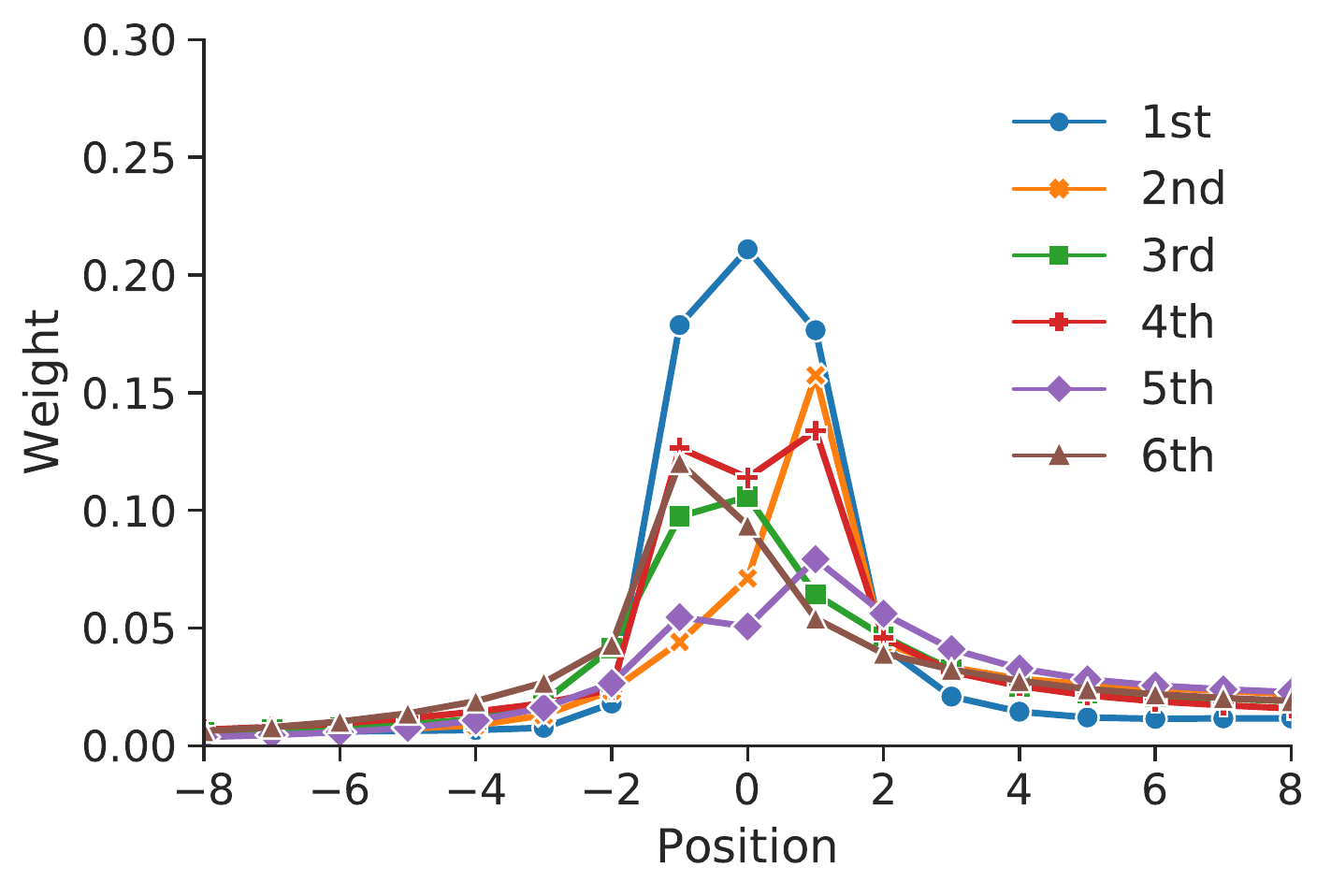}
   \caption{The distributions of self-attentions for each layer on the validation data of the widely used WMT English-to-German dataset. To compute these distributions, we used the encoder side of the 6 layered Transformer encoder-decoder trained on the WMT dataset.}
   \label{fig:attn_distribution}
\end{figure}

\begin{figure*}[!t]
  \centering 
  \includegraphics[width=0.65\textwidth]{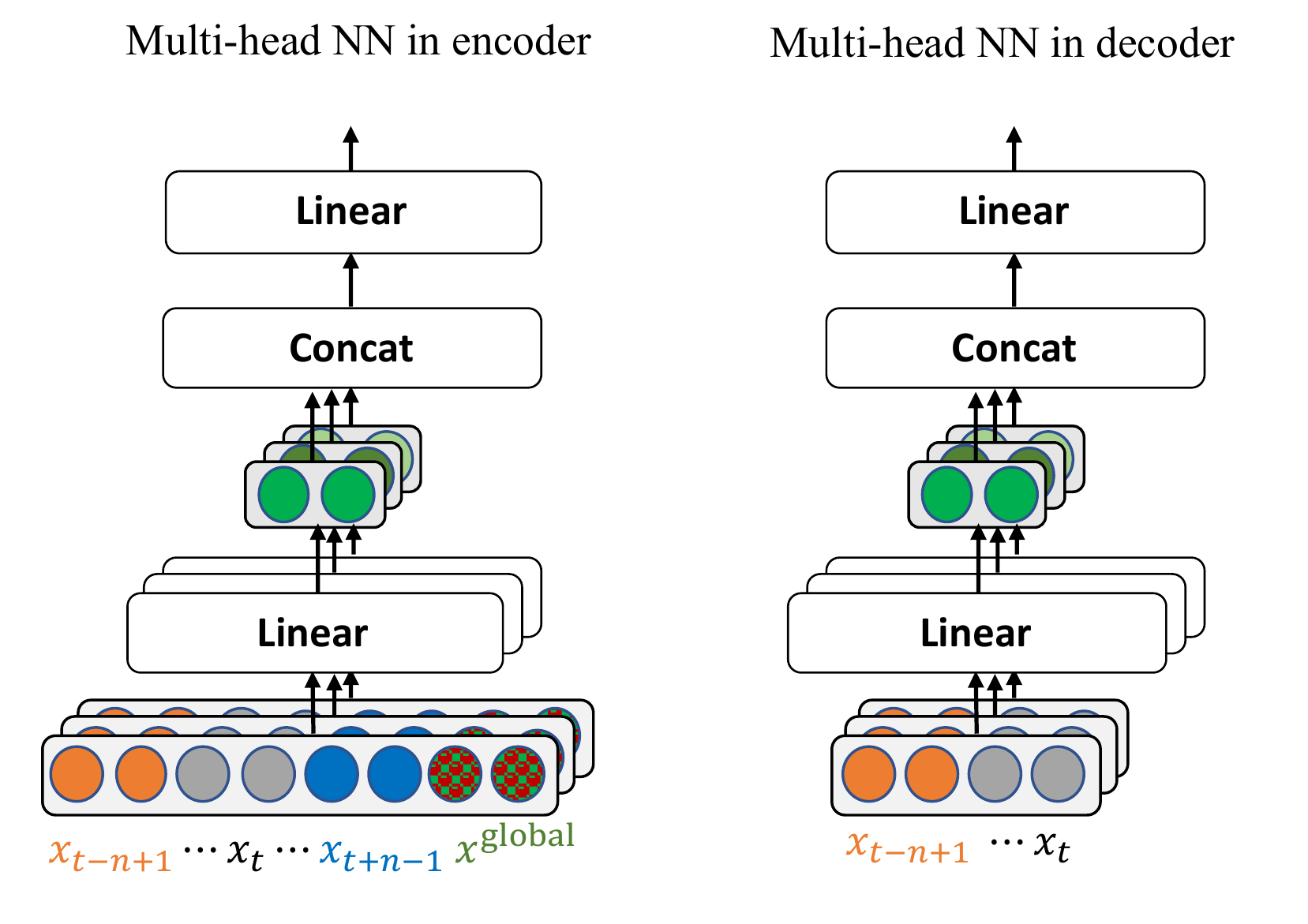}
   \caption{Multi-head neural $n$-gram in encoder (left) and decoder (right).}
   \label{fig:multihead-ngram}
\end{figure*}
However, do we need to consider dependencies among all input representations?
Empirically, self-attentions assign large weights to surrounding several representations.
Figure \ref{fig:attn_distribution} illustrates distributions of self-attentions for each layer on the machine translation task as an example.
Moreover, several studies reported successes of local attentions that restricts the computation of dependencies in the self-attention to surrounding $n$ representations~\cite{sparse-attention-child19,longformer-beltagy}.

For a method using surrounding $n$ representations, an $n$-gram based method is more simple than the self-attention.
\newcite{revisitNgram-sun} significantly improved the performance of neural-based $n$-gram language model~\cite{nplm-begio} by introducing recent techniques such as residual connections~\cite{7780459}.
They reported that their modified method was superior to the traditional neural $n$-gram model~\cite{nplm-begio} but worse than Transformer.
However, do we have no more options to make an $n$-gram based method better than self-attention based method, Transformer?

In this paper, we refine the neural $n$-gram model~\cite{revisitNgram-sun} by introducing the multi-head mechanism used in the multi-head attention in Transformers~\cite{transformer-vaswani}.
We call this reformed $n$-gram model multi-head neural $n$-gram (\textbf{multi-head NN}).
We replace self-attentions in Transformer with this multi-head NN.
We indicate that the proposed method is comparable to or better than Transformer in sequence-to-sequence tasks including machine translation, summarization, and automatic speech recognition (See Appendix \ref{appendix:summarization} and \ref{appendix:asr}).
In addition, we conduct various analyses on the multi-head NN based model to give insights on requirements of computing representations by neural networks.

\section{Multi-Head Self-Attention}
\label{model:transformer}
Multi-head self-attention is a core module of the de facto encoder-decoder, Transformer~\citep{transformer-vaswani}.
Multi-head self-attention captures dependencies between input representations with attention function on entire input representations.
In addition, instead of using single attention function, multi-head self-attention utilizes multiple attention functions on input representations, which are projected with different linear projections.
Specifically, given a sequence of $L$ input representations, $x_1, ..., x_L$, $K$ headed self-attention computes output representation $z_t$ at position $t$ with the following equations:
\begin{align}
    z_t &= \mathrm{concat}(h_{t1}, ..., h_{tK})W \\
    h_{tk} &= \sum_{i=1}^{L}{\alpha_{ti}^{(k)}x_{i}}
\end{align}
, where $\alpha_{ti}^{(k)}$ is scaled dot-product attention weight between input representation at position $t$ and $i$ in $k$-th head introduced in~\newcite{transformer-vaswani}.

\section{Multi-Head Neural $n$-gram}
\label{model:ngram}
We follow architecture of neural $n$-gram language models used in previous works~\cite{nplm-begio, revisitNgram-sun} to build our neural $n$-gram module, but additionally introduce multi-head mechanism as in self-attention of Transformer~\cite{transformer-vaswani}.
Multi-head neural $n$-gram receives concatenation of representation at a specific position and its $n$ surrounding representations as input.
The concatenated representation is passed through multiple linear projections in parallel as in Figure \ref{fig:multihead-ngram}.
Specifically, multi-head NN with $K$ heads computes output representation $z_t$ at position $t$ in sequence of $L$ input representations, $x_1, ..., x_L$, with the following equations.
\begin{align}
    z_t &= \mathrm{concat}(h_{t1}, ..., h_{tK})W \\
    h_{tk} &= \mathrm{ReLU}(\mathrm{concat}(x_{t-n+1}, ..., x_t)W_k)
\end{align}
In encoder, multi-head NN uses bidirectional $n$-gram concatenation to benefit from past and future contexts as model can access to all positions of input sequence at every position.
Specifically, at position $t$, representation from position $t-n+1$ to $t+n-1$ are used to obtain concatenation input.
In additon, we use global context representation as an optional representation in encoder, to expect capturing long-term dependencies between inputs located apart from focused position.
In this paper we compute global context representation by performing max-pooling operation on entire sequence of input representations. 
We simply concatenate this global context representation with $n$-gram concatenation similar to \newcite{revisitNgram-sun}.

\begin{table*}
\centering
\begin{tabular}{lcccc}
\hline
\multirow{2}{*}{Method} & \multicolumn{2}{c}{IWSLT DE-EN} & \multicolumn{2}{c}{WMT EN-DE} \\
 & \#Param. & BLEU & \#Param. & BLEU \\
\hline
Self-attention & 36M & 35.34 & 61M & \textbf{27.20} \\
Local attention & 36M & 34.77 & 61M & 26.71  \\
\hline
\textbf{multi-head NN} & 36M & \textbf{35.49} & 62M & 27.15 \\
\hline
\end{tabular}
\caption{\label{table:main-result-mt}
BLEU scores on the machine translation tasks.
}
\end{table*}

\section{Experiments}
To evaluate performance of multi-head NN, we conducted experiments on machine translation task.
\subsection{Setting}
\paragraph{Datasets}
We experimented on two machine translation benchmarks: WMT English-German (EN-DE) and IWSLT German-English (DE-EN).
For WMT, we used WMT 2016 EN-DE training set~\citep{ott-etal-2018-scaling}, which contains 4.5 million sentence pairs.
We used newstest2013 and newstest2014 as validation and test set respectively in this task.
For IWSLT, we followed the setup in \newcite{edunov-etal-2018-classical} for 160K training sentence pairs.
We used the byte pair encoding~\citep{sennrich-etal-2016-neural} by sharing vocabulary between source and target sides to build vocabulary for each dataset.
The vocabulary size was set to 32K for WMT and 10K for IWSLT.

\paragraph{Methods}
We used Transformer-base~\cite{transformer-vaswani} as a baseline in our experiments.
We compare model performance when replaced multi-head self-attention by multi-head NN.
We adopted local attention~\citep{sparse-attention-child19, longformer-beltagy} as another baseline since it uses only surrounding $n$ representations similar to our multi-head NN.
Note that we remained other components, such as feed-forward networks, layer normalization, and residual connection as in vanilla Transformer.
In local attention and multi-head NN, we set number of surrounding representations to 5 $(n = 5)$ for both encoder and decoder.
For multi-head NN, we use optional global context representation in encoder.
We used 9 layered and 8 layered models when adopting multi-head NN for IWSLT and WMT, respectively, to obtain similar model sizes to Transformer-base.

\begin{table}
\centering
\begin{tabular}{lcc}
\hline
Model & \#Param. & BLEU \\
\hline
Self-attention & 61M & 26.02 \\
Local attention & 61M & 25.50 \\
\cite{revisitNgram-sun} & 62M & 25.41 \\
\hline
multi-head NN 6L& 62M & 25.69 \\
multi-head NN 8L& 62M & 26.00 \\
- multi-head & 62M & 25.33 \\
- global & 62M & 25.80 \\
- global + increasing $n$ & 62M & 26.11 \\
\hline
\end{tabular}
\caption{\label{table:ablation-ngram}
Multi-head NN based model ablations on validation set of WMT.
}
\end{table}

\subsection{Results}
Table \ref{table:main-result-mt} summarizes results on the machine translation tasks.
We report averaged scores over 3 runs with different seeds for each method. 
All results are compared with roughly the same model size.
Multi-head NN achieves comparable BLEU score to self-attention on the WMT dataset.
Interestingly, even with a small margin, multi-head NN outperforms self-attention by $0.15$ BLEU point on the IWSLT dataset.
This result infers that multi-head NN can achieve competitive performance to attention based method without considering dependencies among all input representations at every position.
Moreover, multi-head NN outperforms local attention by more than $0.40$ point, although both methods focus on narrow local context.
This result implies that when restricted context to nearby representations, neural $n$-gram based method with fixed weights can perform better than attention where dynamic weights are computed at every position.

\section{Analysis}
\subsection{Ablation on $n$-gram based Model}
In this section, we evaluate the impact of each component in multi-head NN to give insights on neural $n$-gram based method.
We report results on validation set of WMT dataset.

First, we discuss the impact of model depth on multi-head NN based model.
We hypothesize that increasing model depth helps the multi-head NN broaden its accessible context, contributing to performance improvement.
As in Table \ref{table:ablation-ngram}, with similar number of parameters, a 6 layered multi-head NN underperforms self-attention, but a 8 layered model achieves comparable performance with self-attention.
This result infers that we can improve performance of multi-head NN based method using a deeper model.

Next, we consider effect of multi-head mechanism.
Note that our neural $n$-gram without multi-head mechanism is similar to one in \newcite{revisitNgram-sun}.
Therefore, we also report result when extended neural $n$-gram used in \newcite{revisitNgram-sun} to encoder-decoder setting.
As in Table \ref{table:ablation-ngram}, BLEU scores of multi-head NN without multi-head and one of \newcite{revisitNgram-sun} are comparable but are lower than one of self-attention in a large margin.
This result infers that we need multi-head mechanism to improve performance of neural $n$-gram based model.

\begin{figure}[t]
  \centering 
  \includegraphics[width=8cm]{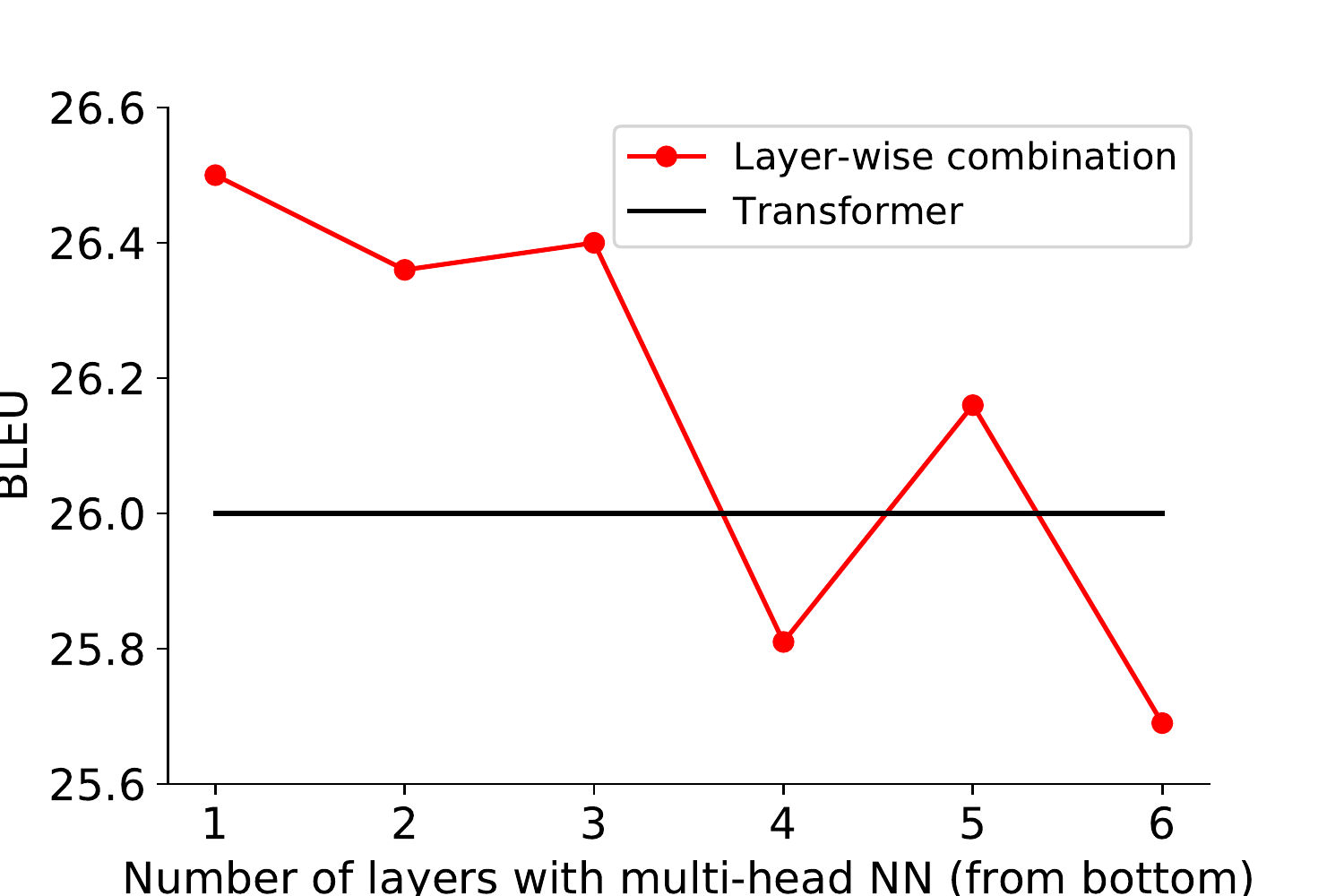}
   \caption{Layer-wise combination of multi-head NN and self-attention.}
   \label{fig:ngram_variant}
\end{figure}

Finally, we discuss effect of global context representation.
As seen in Table \ref{table:ablation-ngram}, removing global context representation from multi-head NN hurts model performance, but in a small margin.
Interestingly, even without global context representation, increasing $n$-gram size as moving to upper layers ($n=[3, 3, 5, 5, 7, 7, 9, 9]$) can also achieve competitive performance.
Performance in this setting is even better than the one with global context representation.
This result infers that incrementally increasing context range across layers of multi-head NN based model tends to improve the performance.

\subsection{Better to Hybridize $n$-gram and Attention}
Inspired by the competitive performance of multi-head NN even with limited context, we investigate combinations of multi-head NN with multi-head self-attention.
We expect to improve encoder-decoder performance as combinations of both modules would benefit from different types of dependency capturing.
We consider two combinations: layer-wise and encoder/decoder combinations.

Figure \ref{fig:ngram_variant} shows BLEU score on validation set of WMT dataset when hybridized multi-head NN without global context and self-attention across layers of 6 layered Transformer.
As in Figure \ref{fig:ngram_variant}, using multi-head NN in first three layers improve BLEU score up to $0.50$ point over vanilla Transformer\footnote{This result is statistically significant according to Student’s t-test (p < 0.05)}.
However, this improvement decreases when number of layers with multi-head NN is increased.
This result suggests the importance of a narrow context focusing at lower layers.

\begin{table}
\centering
\begin{tabular}{llc}
\hline
Encoder & Decoder &  BLEU \\
\hline
Self-attention & Self-attention  & 26.02 \\
Self-attention & multi-head NN & 26.46 \\
multi-head NN & multi-head NN  & 26.00 \\
multi-head NN & Self-attention & 25.95 \\
\hline
\end{tabular}
\caption{\label{table:ablation-encdec}
Combination of multi-head NN and self-attention in encoder and decoder.
}
\end{table}

Table \ref{table:ablation-encdec} shows BLEU score on validation set of WMT dataset when hybridized multi-head NN and self-attention in encoder and decoder.
Replacing multi-head self-attention in encoder with multi-head NN tends to hurt performance, although with slight drops compared to vanilla Transformer.
However, replacing only multi-head self-attention in decoder with multi-head NN improves performance by a large margin over vanilla Transformer\footnotemark[\value{footnote}].

Results from these two combination variants infer that multi-head NN is complementary to multi-head self-attention.
We suggest using proper combinations of multi-head NN and self-attention to improve performance of encoder-decoder.

\section{Related work}
Regarding necessary of context to attend in attention based methods, recent works found that variants of self-attention with constricted context can also achieve competitive performance~\cite{sparse-attention-child19, longformer-beltagy}.
In this work, we also work on limited context representation, but instead of using attention based methods, we focus on feed-forward neural networks, which is more straightforward.
Our work is based on neural $n$-gram model, which is close to \newcite{revisitNgram-sun}, but we further improved performance by adopting multi-head mechanism as in \newcite{transformer-vaswani}.

\section{Conclusion}
In this paper, we introduced multi-head mechanism to neural $n$-gram model to improve performance on sequence modeling.
We showed competitive performance of multi-head NN based method compared to attention based method in sequence-to-sequence tasks, although context representation is constricted to a narrow range.
Our analyses found that multi-head NN and self-attention are complementary to each other as their combinations can even outperform vanilla Transformer.

\bibliography{ngram}
\bibliographystyle{acl_natbib}

\newpage
\appendix

\section{Text Summarization}
\label{appendix:summarization}
\paragraph{Setting}
We experimented with two summarization tasks: headline generation and document summarization.
The headline generation is a sentence-level summarization task in which model is asked to generate a corresponding headline given a sentence of news article.
We used Gigaword dataset~\citep{rush-etal-2015-neural} in this task.
The document summarization is a document-level summarization task in which the model is asked to generate a summary consisting of several sentences from a long document.
We used CNN/DailyMail dataset~\citep{see-etal-2017-get} in this task.
We used source-target joined vocabulary constructed using the byte pair encoding~\citep{sennrich-etal-2016-neural} with size of 32K in both tasks.

\paragraph{Result}
Table \ref{table:main-result-sum} shows results on the headline generation and document summarization tasks.
We report results under the same model size setting as in the machine translation experiment.
In both the headline generation and document summarization tasks, multi-head NN achieves competitive performance compared to self-attention.
Moreover, while similarly focusing on only surrounding context representations, multi-head NN outperforms local attention with large margins.
This trend is similar to ones in the machine translation tasks.

\begin{table*}
\centering
\begin{tabular}{lccccccc}
\hline
\multirow{2}{*}{Method} & \multirow{2}{*}{\#Param.} & \multicolumn{3}{c}{Headline Generation} & \multicolumn{3}{c}{Document Summarization} \\
  & & R-1 & R-2 & R-L & R-1 & R-2 & R-L \\
\hline
Self-attention & 61M & \textbf{37.80} & 18.91 & 34.91  & \textbf{39.15} & 17.21 & \textbf{36.43} \\
Local attention & 61M & 34.72 & 17.34 & 31.75 & 34.27 & 14.57 & 32.24 \\
\hline
\textbf{multi-head NN} & 62M & 37.71 & \textbf{19.02} & \textbf{35.17}  & 39.04 & \textbf{17.38} & 36.01 \\
\hline
\end{tabular}
\caption{\label{table:main-result-sum}
ROUGE-1, 2 and L (R-1, 2 and L) F1 scores on the summarization tasks.
}
\end{table*}

\section{Speech Recognition}
\label{appendix:asr}
\paragraph{Setting}
Besides text-to-text tasks, we experimented with speech-to-text task with LibriSpeech dataset~\citep{librispeech}.
LibriSpeech contains approximately 960 hours of speech from audiobooks.
There are two data pools, clean and other, in this dataset based on word error rate of speakers.
We used both pools in our experiment.
Vocabulary for decoder side was constructed using SentencePiece~\citep{kudo-richardson-2018-sentencepiece} with a vocabulary size of 10K.

\paragraph{Result}
Table \ref{table:main-result-s2t} shows results on the speech recognition task in development and test set of LibriSpeech dataset.
We report word error rate in this table.
As seen in Table \ref{table:main-result-s2t}, multi-head NN performs on-par with self-attention with approximately the same model size.
Moreover, multi-head NN achieves lower word error rate than local attention by more than 1 point. 
These results are also consistent with ones on the machine translation and text summarization tasks.

\begin{table*}
\centering
\begin{tabular}{lcrrrr}
\hline
 Method & \#Param. & dev-clean & dev-other & test-clean & test-other \\
\hline
Self-attention & 52M & 3.58 & 8.86 & 4.06 & \textbf{8.67} \\
Local attention & 52M & 5.91 & 10.02 & 6.43 & 10.46 \\
\hline
\textbf{multi-head NN} & 52M & \textbf{3.27} & \textbf{8.83} & \textbf{3.52} & 8.70 \\
\hline
\end{tabular}
\caption{\label{table:main-result-s2t}
Word error rate on the speech recognition task.
}
\end{table*}

\end{document}